\documentclass[a4paper]{article}
\usepackage{proceedings}
\usepackage{times}
\usepackage[english]{babel}  
\usepackage{graphicx}
\usepackage{latexsym}

\title{Linking Makinson and Kraus-Lehmann-Magidor preferential entailments}

\author{{\bf Yves Moinard}\\
IRISA, Campus de Beaulieu,\\
35042 RENNES-Cedex FRANCE,\\
tel.:~(33)  2 99 84 73 13,\\
e-mail:~moinard@irisa.fr}

\begin{document}
\maketitle
\bibliographystyle{plain}

\thispagestyle{empty}
\begin{abstract}
  About ten years ago, various  notions of preferential entailment have been
  introduced. 
 The main reference is a paper by 
Kraus, Lehmann and Magidor (KLM), one of the
main competitor being a
more general version defined by Makinson (MAK). 
These two versions have already been compared, but 
it is time to revisit these 
comparisons. Here are our three main results:
(1)  These two notions are
equivalent, provided that we restrict our attention, as done in KLM,
to the cases where the entailment respects logical equivalence (on the left
and on the right).
(2) A serious simplification of the description of the fundamental cases
in which MAK is equivalent to KLM, including a natural passage
in both ways. 
(3) The two previous results are given for preferential entailments more
general 
than considered in some of the original texts, but they apply
also to the original definitions and, for this particular case also, the
models can be simplified. 
\end{abstract}
\newcommand{\qq}{\forall}
\newcommand{\ie}{\exists}
\newcommand{\imp}{\Rightarrow}
\newcommand{\eq}{\Leftrightarrow}
\newcommand{\et}{\wedge}
\newcommand{\ou}{\vee}
\newcommand{\inc}{\subset}
\newcommand{\incl}{\subseteq}
\newcommand{\non}{\neg}
\newcommand{\ent}{\vdash}
\newcommand{\mod}{\models}  
\newcommand{\trimod}{\makebox[1em]{$|\hspace{-.42em}\equiv$}} 
\newcommand{\trimodi}{|\!\equiv}  
\newcommand{\inter}{\cap}
\newcommand{\union}{\cup}
\newcommand{\vide}{\emptyset}
\newcommand{\fleche}{\rightarrow}
\renewcommand{\sup}{\sqcup}
\newcommand{\bsup}{\bigsqcup}
\newcommand{\entnm}{\mid \hspace{-.2em} \sim}

\newcommand{\C}{\mbox{$\cal C$}}
\newcommand{\La}{\mbox{\bf L}}
\newcommand{\MM}{\mbox{\bf M}}
\newcommand{\MMi}{{\bf M}} 
\renewcommand{\P}{\mbox{$\cal P$}}
\renewcommand{\SS}{\mbox{\bf S}}
\newcommand{\T}{\mbox{$\cal T$}}
\newcommand{\Ti}{{\cal T}}  
\newcommand{\iT}{\mbox{$_{\cal T}$}} 
\newcommand{\TT}{\mbox{\bf T}}
\newcommand{\WW}{\mbox{\bf W}}
\newtheorem{theo}{Theorem}[section]
\newtheorem{ex}[theo]{Example}
\newtheorem{ex_0}{Example}[section]
\newtheorem{defi}[theo]{Definition} 
\newtheorem{defis}[theo]{Definitions}
\newtheorem{notat}[theo]{Notation}
\newtheorem{notats}[theo]{Notations}
\newtheorem{cor}[theo]{Corollary}
\newtheorem{prop}[theo]{Property}
\newtheorem{propo}[theo]{Proposition}
\newtheorem{lem}[theo]{Lemma}
\newtheorem{rem}[theo]{Remark}
\newtheorem{rems}[theo]{Remarks} 
\newtheorem{com}[theo]{Comment}
\newtheorem{coms}[theo]{Comments} 
\newtheorem{conj}[theo]{Conjecture}
\newtheorem{theo2}{Theorem}[subsection]
\newtheorem{ex2}[theo2]{Example}
\newtheorem{ex2_0}{Example}[subsection]
\newtheorem{defi2}[theo2]{Definition} 
\newtheorem{defis2}[theo2]{Definitions}
\newtheorem{notat2}[theo2]{Notation}
\newtheorem{notats2}[theo2]{Notations}
\newtheorem{cor2}[theo2]{Corollary}
\newtheorem{prop2}[theo2]{Property}
\newtheorem{propo2}[theo2]{Proposition}
\newtheorem{lem2}[theo2]{Lemma}
\newtheorem{rem2}[theo2]{Remark}
\newtheorem{rems2}[theo2]{Remarks} 
\newtheorem{com2}[theo2]{Comment}
\newtheorem{coms2}[theo2]{Comments} 
\newtheorem{conj2}[theo2]{Conjecture}

\pagestyle{empty}
\section{INTRODUCTION}
Here is one possible presentation of preferential entailments:
We are given some 
knowledge, 
represented as a set of logical formulas.
This set 
can be associated with various kinds of objects, providing its ``semantics'':
it can be associated with its set of models, 
or equivalently in the 
propositional case, with the set of the complete theories which entail
the formulas. 
Or it can be associated with
the set of theories (complete or not) which entail the 
formulas. 
Then we are given a binary relation among these objects, 
and we keep only the objects which are
``preferred'' (meaning minimal) for this relation. 
We get 
a stronger set of formulas, 
deduced ``by default'': 
we get also all the formulas associated with this reduced set of 
objects. This allows to reason in a non monotonic way, since 
augmenting the 
knowledge may invalidate previous
conclusions. Indeed, some objects 
may become minimal in the smaller set associated with the new 
knowledge.
We can allow more flexibility by considering copies of models, or copies 
of theories, 
defining the relation
among these sets of copies. We get then four kinds of preferential 
entailments, called KLM 
 below,
which have been introduced by Kraus, Lehmann and Magidor (1990) in \cite{KLM90}
(\cite{KLM90} requires some conditions on the relation, but 
adding these conditions is straightforward in our results).

Makinson (1994, first version in 1989) has defined a more general version
\cite{Mak94G}, called MAK here. 
An unstructured ``semantics'' is defined simply by a satisfaction
relation from 
some set of objects to the set of formulas, without any condition.
It is then useless to consider sets of these objects instead of singletons:
its suffices to define directly as our starting set, the set of sets 
that we would want to consider. Also, since nothing prevents two different
objects from being associated with the same set of formulas, it is useless to
consider copies of objects. 
A drawback of this simple definition is that the notion of classical deduction
is 
lost. The ``entailment'' can be highly non standard, departing from classical
logic and natural ways of reasoning.
We can e.g. deduce $A\et B$ without deducing $A$.
As a bonus, we can consider as our starting logic basically any non
classical logic. 

If we want to compare the two notions, we must determine which
non standard behavior we admit in the ``semantics''
defining the preferential entailment.
If we want to extend KLM entailment in order to deal with any situation
accessible to MAK entailment, all we have to do is to admit 
any unconstrained relation $\trimod$ between the set of states and the set of
formulas. 
We can even restrict our attention to the simplest of the
four cases, defining the relation directly on the set of the objects 
describing the semantics.
Thus, the interesting point is the other direction.
We show that, provided classical equivalence is respected, MAK
entailment  
is equivalent to KLM entailment. We describe a simple 
subclass of MAK entailment, 
which includes all the  cases where 
MAK entailment respects classical equivalence, and for which it is easy to
describe a 
passage from MAK formulation to KLM formulation and
back. 
We improve previous results obtained by Dix and Makinson (1992) in 
\cite{DM92} and by Voorbraak (1993) in \cite{Voo93P}. With respect to
\cite{DM92}, the   
description of the subclass of MAK is much simpler. 
Thanks to recent results on preferential entailments, we establish
equivalence between the two formalisms in all the cases where it is possible, 
namely when MAK respects classical equivalence. 

In Section \ref{secnot} 
we introduce the notations and the logical pre-requisite necessary for 
this text. 
In Section  \ref{secdef} we remind the
definitions and main properties of the kind of preferential entailments
considered here, giving our results
in Section  \ref{secrel}.

\section{NOTATIONS AND FRAMEWORK}\label{secnot}
$\bullet$ $\La, \; \varphi, \; \T$:
We work in a propositional language $\La$, 
and we use the same denotation $\La$ for its set of formulas.
Letters $\varphi, \psi$ denote {\it formulas}
(identified with their equivalence class).
Letters $\T$ or $\C$ denote sets of formulas.

$\bullet$ $V, \; \MM, \; \mu, \; \P(E), \; \mu \mod \cdots$:
$V(\La)$ (vocabulary), denotes a set of  propositional symbols
and $\mu$ 
denotes an {\it interpretation for $\La$},
identified with the subset of $V(\La)$ that it satisfies.
The {\it satisfaction relation} is denoted by $\mod$, $\mu \mod \varphi$ and
$\mu \mod \T$ being defined classically. 
For any set $E$, $\P(E)$ denotes the set of its subsets.
The set $\P(V(\La))$ of the interpretations for $\La$ is denoted by $\MM$.
A {\it model} of $\T$ is an interpretation $\mu$ such that $\mu \mod \T$.
The sets of the models of $\T$ and $\varphi$ are denoted respectively by 
$\MM(\T)$ and $\MM(\varphi)$.

$\bullet$ $\T \mod \cdots, Th(\T), \TT$:
$\T \mod \varphi$ and $\T \mod \T_1$ are defined classically.
A {\it theory} is a subset of $\La$ closed for deduction, and
$\TT$ denotes the set $\{\T\incl \La\; /\; \T=Th(\T)\}$
of the theories of $\La$.

$\bullet$ $\MM_1 \mod \cdots, Th(\mu), Th(\MM_1)$:
A theory $\C\in \TT$ is {\it complete} if 
$\qq \varphi \in \La$, $\varphi \in \C$ iff $\non \varphi \notin \C$.
The set $Th(\mu) = \{\varphi \! \in \!\La \; /\; \mu \!\mod \!\varphi\}$ 
of the formulas satisfied by $\mu$ is the {\it theory of $\mu$}. 
For any subset $\MM_1$ of $\MM$, 
$\MM_1 \mod \T$ means 
$\mu \mod \T$ for any $\mu\in \MM_1$ and the {\it theory of $\MM_1$} is the set
$Th(\MM_1)=\{\varphi \in \La\;/\; \MM_1 \mod \varphi\}$ [thus $Th(\MM_1)=
\bigcap_{\mu \in \MMi_1} Th(\mu)$].
This ambiguous use of $Th$ and of $\mod$
(applied to sets of formulas or interpretations) is usual. 
For any $\T \in \TT$, we get 
$\T=\bigcap_{{\cal T}_i \in 
 {\bf T},\; {\cal T}_i \mod {\cal T}} \;\T_i = \bigcap_{{\cal C}_i \in 
 {\bf T},\; {\cal C}_i {\rm \ complete},\;{\cal C}_i \mod {\cal T}} \;\C_i$.
The set of all the complete theories
is in a natural one-to-one mapping with $\MM$:
For any $\mu \in \MM$, $Th(\mu)$ is complete, and for any complete $\C\in\TT$,
$\MM(\C)$ is a singleton $\{\mu\}\incl \MM$. For any $\T \incl \La$, we get
$\mu \mod \T$ iff $Th(\mu) \mod \T$ and for any $\MM_1 \incl \MM$,
$\MM_1 \mod \T$ iff $Th(\MM_1) \mod \T$.

\section{PREFERENTIAL ENTAILMENTS}\label{secdef}
\subsection{PREFERENTIAL KLM ENTAILMENT}
Since their introduction \cite{KLM90}, these kinds of preferential
entailment have been extensively studied. As \cite{DM92} remarks, 
``the use of the term {\em preferential} is [..] 
rather anarchic [...]''. The situation has not really improved
since these ``early years'', however, it is clear that now the word 
is not restricted to the ``cumulative cases'' as done in \cite{KLM90}.
The expression ``preferential entailment'' was first introduced by Shoham 
(1988)
in \cite{Sho88}, and then regularly generalized and/or modified.
The basic idea however is still the same: we consider a set of objects
describing the semantics, and a binary relation $\prec$ on this set of objects.
We get a ``preferential semantics'' in which only the objects,
associated with a set of formulas, which are minimal for $\prec$, are
considered. 
The definitions we give can be found in e.g. \cite[Definitions 3.10,
3.13]{KLM90} (``single formula version'') 
and \cite[Definitions 4.26--29]{FL93}
(``theory version'', only version considered here), 
with some modifications which have already been considered in e.g. 
\cite{DM92,Boc99C,Eng00}. These modifications are either cosmetic, or consist
in dropping 
some special condition imposed in 
the original text to the relation $\prec$, since (1) we do not need these
restrictions, and (2) our study can accommodate in a straightforward way 
these restrictions.

\begin{defi}\label{defklmmod} 
A {\em KLM model} is a triple $\SS=(S,l,\prec)$, 
where $S$ is a set, the
  elements of which are called {\em states}, $l$ is a mapping
$S\rightarrow \P(\MM)$ that labels every state with a set of
interpretations and $\prec$ is a binary relation on $S$, called 
a {\em preference relation}.

We define a {\em satisfaction relation} $\trimod$: for any $s \in S$,
$s \trimod \varphi$ whenever $l(s) \mod \varphi$ and,
for any $\T \incl \La$,
$s \trimod \T$ whenever $l(s) \mod \T$.
For any set of formulas $\T \incl \La$, we define 
$S(\T)= \{s\in \SS\;  / \; s \trimod \T\}$ and the set $S_\prec(\T)$ of the
states in $S(\T)$ which are {\em minimal for $\prec$} by
$S_\prec(\T)= \{s\in S(\T)\; / \;s'\prec s \mbox{ \ for no \ } s'\in S(\T)\}$.
\end{defi}
$S(\T)$ is $\widehat{\T}$ in the original texts.
Notice that, as noted by Bochman (1999) in \cite{Boc99U}, we can replace the
set $l(s)$ of 
interpretations by a theory, precisely the theory $Th(l(s))$,
and we will in fact generally prefer this formulation, where $l$ is a mapping
$S\rightarrow\TT$ instead of $S\rightarrow\P(\MM)$.
Also, we drop here the {\em consistency of states condition} 
$l(s) \not = \vide$ (or alternatively
$l(s) \not = \La$ if we consider labelling with theories) which appears in the
original definitions. As explained below, this condition is unnecessary. 

The role of $l$ is to allow {\em ``copies of'' sets of interpretations}
(or alternatively {\em ``copies'' of theories}), since various states can be
mapped by $l$ to the same object. 

\begin{defi} \label{defent}
Let us call an {\em entailment relation}, any relation $\entnm$  in
$\P(\La)\times\La$.
Any entailment relation can be extended into a relation in
$\P(\La)\times\P(\La)$ by defining  $\T \entnm \T'$ as $\T \entnm \varphi$ for
any $\varphi\in \T'$.

From any entailment relation, we can define a mapping $C$ from $\P(\La)$ to
itself, called an {\em entailment}, as follows: 
$C(\T) = \{\varphi \in \La\;/ \;\T\entnm \varphi\}$.
\end{defi}

\begin{defi} \label{defklmgpe}\cite{KLM90}
A {\em KLM entailment relation} $\entnm_\prec^{KLM}$ is defined as
  follows from a KLM model $\SS$: for any $\T\union\{\varphi\}\incl \La$,
$\T \entnm_\prec^{KLM} \varphi$ whenever $s\trimod \varphi$ for any 
$s \in S_\prec(\T)$. We write also $\varphi\in C_{KLM}(\T)$ 
instead of $\T \entnm_\prec^{KLM} \varphi$ and call the entailment $C_{KLM}$ 
 a {\em KLM preferential entailment}, or
a {\em KLM entailment} for short.
\end{defi}

\begin{defi} \label{defprecirc}
A {\em pre-circumscription} $f$ (in \La) is an extensive 
(i.e., $f(\T) \supseteq \T$ for any $\T$) mapping from $\TT$ to $\TT$.
For any subset $\T$ of \La, 
we use the abbreviation $f(\T) = f(Th(\T))$, assimilating a 
pre-circumscription to a
particular extensive entailment.
We write $f(\varphi)$ for $f(\{\varphi\})=f(Th(\varphi))$.
\end{defi}

Thus, we call here pre-circumscription any entailment 
which respects full logical equivalence
and which is extensive. By ``respects full logical equivalence'', we mean
that, if $\T_1$ and $\T_2$ are two logically
equivalent sets of formulas [i.e. $Th(\T_1) = Th(\T_2)$], then 
(1) (``left side'') $f(\T_1) = f(\T_2)$, and 
(2) (``right side'') 
$\T_1\incl f(\T)$ iff $\T_2 \incl f(\T)$.
The ``right side'' is equivalent to ``right weakening'':
if $\T_1 \mod \varphi$ and $\T_1\incl f(\T)$, then $\varphi \in f(\T)$.

\begin{defi} \label{defct}
An entailment $C$ satisfies  (CT),
{\em cumulative transitivity}, also known as ``cut'', if
\[
\mbox{for any }\T' \incl C(\T), \mbox{ we get }
 C(\T \union \T') \incl C(\T).\] 
\end{defi}

Here are the two main (and characterizing) properties of KLM entailments:

\begin{prop}\label{propklmprecircct} \cite{KLM90,Mak94G}
Any KLM entailment $C_{KLM}$ is a pre-circumscription  
satisfying {\rm (CT)}. 
\end{prop}
\begin{prop}\label{propklmcar} \cite{Voo93P,Moi00G}
Any pre-circumscription satisfying {\rm (CT)} is a KLM entailment. 
\end{prop}

Particular KLM models can be considered. The three kinds described
now originate also from \cite{KLM90}, where no special names are given.
Let $\SS=(S, l, \prec)$ be a KLM model.
\begin{defi}\label{defpartklmmod}
  \begin{enumerate}
  \item If $S=\P(\MM)$ (or equivalently, under the alternative formulation in
    terms of theories, $S=\TT$) and $l = identity$, then $\SS$ is a {\em
      simplified} (or {\em unlabelled}) KLM model. 
  \item If each $l(s)$ is a singleton in $\P(\MM)$ (or equivalently a complete
    theory), then $\SS$ is a {\em singular} KLM model. 
  \item If $\SS$ is simplified and singular, then $\SS$ is a {\em strictly
      singular} KLM model.  
  \end{enumerate}
\end{defi}
With the unrestricted case, we get then four kinds of models, which could give
rise to four kinds of KLM entailments.
It happens (\cite{Moi00G}, this result could also be extracted from an
independent work by Voorbraak \cite{Voo93P}) that we can without lack of
generality restrict our attention to simplified KLM models: in the proof of
Property \ref{propklmcar}, we can easily get a simplified KLM model.
Thus, we do not really need to use ``states'' in KLM models.
Notice that in the particular case
of a singular KLM model, we generally cannot  
suppress the states if we want to keep only singletons in the image $l(S)$ of
$l$ (as shown in a very simple finite example in \cite[p.193]{KLM90}, 
which applies  
here): we cannot suppress the states without
leaving this attractive particular case\footnote{%
Even in this case, the states can be suppressed, provided we enlarge the
vocabulary of the initial language $\La$ in such a way that each different
state gives rise to a different interpretation in the new language: this
method is introduced by Costello (1998) in \cite{Cos98} for the cumulative and
finite case}.   
This means that if we start from a preference relation defined in a set of
copies of interpretations, we cannot always get an equivalent relation defined
directly  
on the set $\MM$ of the interpretations. This feature is a good motivation for
using states, but only in the case of singular models.   
Also we cannot express any KLM  
entailment thanks to a singular KLM model (a
small finite counter-example in \cite[proof of Lemma 4.5]{KLM90} applies 
here)\footnote{%
Again, this limitation can be overcome \cite{Moi01}, at least in the finite
case: at the price of a severe modification of the vocabulary of the initial
language $\La$, 
any KLM preferential entailment can be expressed in terms of a strictly
singular KLM model.}. This means that if we start from a preference relation
defined on  
a set of copies of sets of interpretations (or equivalently of copies of
theories), then we 
can find an equivalent relation defined directly on the set $\cal{P}(\MM)$ 
(or $\TT$), but we generally cannot define the relation on the set $\MM$
of interpretations or even on a set of copies of interpretations.
 Thus, we get exactly three kinds of KLM preferential
entailments (see their syntactical characterizations in \cite{MR02C}), instead
of four. A consequence of this reduction to simplified 
models is that any singular model is equivalent to a simplified model, which
is not absolutely obvious from the definitions
(``equivalent'', meaning here giving rise to the
same preferential entailment).

\subsection{PREFERENTIAL MAK  ENTAILMENT}

Makinson considers an entailment more general than its KLM
counterpart, with a simpler definition. 
The price is that this notion leaves classical consequence
altogether, getting a highly non standard preferential entailment in which 
we can conclude $A \et B$ without concluding $B$, and in which
what we conclude from $A \et B$ is not related to what we conclude 
from $\{A, B\}$. This can be useful if we want to extend the notion of
preferential entailment to non classical logics. However, if we want to stay
in  
our good old classical way of reasoning, this is rather confusing. In any
case, a fair comparison with KLM definitions needs to equate the ways we want
to reason at first. 
This is what we will do in Section \ref{secrel}, 
after giving the definitions now.

\begin{defi}\label{defmakmod}\cite{Mak94G}
A {\em MAK model} is a triple \mbox{$\SS=(S,\trimod, \prec)$} where 
$S$ is a set, the elements of which are called {\em states}, 
$\prec$ is a binary relation on $S$, called a {\em preference relation}
(till now, this is as in KLM models Definition \ref{defklmmod})
and where $\trimod$ is any {\em satisfaction relation} on $S$.
We write $s \trimod_\prec \varphi$ [respectively $s \trimod_\prec \T$]
whenever $s \trimod \varphi$ [respectively $s \trimod \T$, i.e. 
$s \trimod \varphi$ for any $\varphi\in\T$]
and for no $s' \in S$ such that $s'\prec s$ we have $s' \trimod \varphi$
[respectively $s' \trimod \T$].

A {\em MAK entailment relation} $\entnm_\prec^{MAK}$ is defined as follows:
For any set of formulas $\T \union\{\varphi\}$, 
$\T \entnm_\prec^{MAK} \varphi$ whenever $s \trimod \varphi$ for all $s\in S$
satisfying $s \trimod_\prec \T$.

A {\em MAK preferential entailment}, or {\em MAK entailment} for
short, is an entailment  $C_{MAK}$
defined by a MAK entailment relation:
$C_{MAK}(\T)=\{\varphi\in\La \;/ \;\T \entnm_\prec^{MAK} \varphi\}$.
\end{defi}
The names KLM model and MAK model are from \cite{DM92}.
What makes this short definition so powerful is 
that no condition is required for $\trimod$.
This makes the preferential entailment very different from what we
expect for an ``entailment''. 
$C_{MAK}$ is far from being a pre-circumscription:
if $\T_1$ and $\T_2$ are classically equivalent, we do not know anything about
$C_{MAK}(\T_2)$ when we know $C_{MAK}(\T_1)$. Moreover, we almost need an
extensive description of all the sets $C_{MAK}(\T)$, since they are
not classical theories. If we drop the identification between a formula and
its equivalence class, we can even consider logics where $C_{MAK}(A \et B)$ is
different from $C_{MAK}(B \et A)$.
It may even seem strange that such a formalism has interesting
properties, however we get cumulative transitivity:

\begin{prop}\label{propmakct}\cite{Mak94G}
Any MAK  entailment is extensive and satisfies (CT).
\end{prop}

This implies also {\em idempotence} (Makinson, \cite{Mak94G}):\\
$C_{MAK}(C_{MAK}(\T))=C_{MAK}(\T)$. 
Nevertheless, the significance of (CT)
for such a non standard ``entailment'' is  
far from being as great as when we deal only with pre-circumscriptions.
 
Before making a comparison between MAK and KLM, let us give a few
natural definitions, which extend to ($S$, $\trimod$) 
what is usually
done with classical interpretations ($\MM$, $\mod$).

\begin{defis}\label{defsmak}
Let $(S,\trimod, \prec)$ be a MAK model. 
As in Definition \ref{defklmmod},  for $s\in S$, $\T\incl \La$, and
$\varphi\in\La$, $\;s \trimod \T$ means $s \trimod \varphi$ for any
$\varphi \!\in \!\T$,  
$S(\T) = \{ s \!\in  \!S / s \trimod \T\}$ {\ \em [and }
$S(\varphi) = S(\{\varphi\})$%
{\em]}.\\
We define the entailment $Cn_{\trimodi}$ as follows: 
$Cn_{\trimodi}(\T) = \{ \varphi\in\La /\mbox{ for any } s \!\in \!S(\T),
 s \trimod \varphi\}$.

We define also, for each $s\in S$: 
 $Cn_{\trimodi}(s) = \{\varphi\in\La \,/\, s \trimod \varphi\}$.
 \end{defis}
We get (straightforward \cite{DM92}):
$Cn_{\trimodi}$ is a {\em Tarski entailment},
i.e. it is
an extensive entailment satisfying idempotence (point \ref{lab1}) and {\em
  monotony} (point \ref{lab2}):
for any sets $\T, \T'$ of formulas,\vspace{-1ex}
\begin{enumerate}
\item\label{lab1}
$\T \incl Cn_{\trimodi}(\T) = Cn_{\trimodi}(Cn_{\trimodi}(\T))$.\vspace{-1ex}
\item\label{lab2} 
If $\T \incl \T'$ then $Cn_{\trimodi}(\T)\incl Cn_{\trimodi}(\T')$.
\end{enumerate}
Notice that we get (immediate from the definitions):
\hspace{1em}
$Cn_{\trimodi}(\T) = \bigcap_{s \in S(\Ti)} Cn_{\trimodi}(s)$.\\
As a particular case, we get $Cn_{\trimodi}(\T) = \La$ if 
$S(\T)=\vide$.

As in classical logic, we use the same notation $Cn_{\trimodi}$
for two different, but closely related, notions (cf $Th$ in Section
\ref{secnot}): 
an entailment, defined by $Cn_{\trimodi}(\T)$, 
and the notion of ``theory of a state'', defined by $Cn_{\trimodi}(s)$.
A justification for using the same writing $Cn_{\trimodi}$ 
is that $Cn_{\trimodi}(s)$ is indeed a {\em theory in the meaning of
  $Cn_{\trimodi}$}: 
 for any $s \in S$, $Cn_{\trimodi}(Cn_{\trimodi}(s)) = Cn_{\trimodi}(s)$.

\underline{Proof:} $Cn_{\trimodi}(s)\incl Cn_{\trimodi}(Cn_{\trimodi}(s))$ 
since $Cn_{\trimodi}$ is a Tarski entailment.\\
$Cn_{\trimodi}(Cn_{\trimodi}(s))$ $ =$ $ \bigcap_{s'\in
  S(Cn_{\trimodi}(s))} Cn_{\trimodi}(s')$ by definition of
 $Cn_{\trimodi}(\T)$,
and $s \in S(Cn_{\trimodi}(s))$ [i.e. $s \trimod Cn_{\trimodi}(s)$], thus
$Cn_{\trimodi}(Cn_{\trimodi}(s))\incl Cn_{\trimodi}(s)$.

Here is a last result of this kind \cite{DM92}:
\hspace{1em}
For any $\T\incl \La$, we get $Cn_{\trimodi}(\T) \incl C_{MAK}(\T)$.

Notice that, contrarily to $Cn_{\trimodi}$, 
 $C_{MAK}$ is not a Tarski entailment: it falsifies monotony.

\section{RELATING MAK AND KLM ENTAILMENTS}\label{secrel}

\subsection{THE MAK ENTAILMENTS WHICH ARE KLM ENTAILMENTS}
MAK notion encompasses KLM notion, as noticed in
\cite{Mak94G,DM92}:

\begin{prop}\label{propklmmmak} 
For each KLM model $\SS = (S, l, \prec)$ (defining a KLM
entailment $C_{KLM}$),
there exists a MAK model $(S, \trimod, \prec)$
such that its associated MAK entailment $C_{MAK}$ is equal to
$C_{KLM}$.
\end{prop}
Notice that  we even do not need to make $l$
to vanish in the KLM model (even if we know that we could do so without loss 
of generality).
The set of states and the preference relation are unmodified.
It suffices to define $\trimod$, as in \cite{DM92}, by:
$s \trimod \varphi$ whenever $l(s) \mod \varphi$.
It is clear from the definitions that we get indeed 
$\entnm_\prec^{MAK} = \entnm_\prec^{KLM}$. 

What is interesting then is to describe precisely the subclass of 
MAK models which can be translated into KLM models.
Here is a characterization of this subclass:

\begin{theo}\label{thmakprecirc}
A MAK model \mbox{$\SS = (S, \trimod, \prec)$}
gives rise to a MAK  entailment $C_{MAK}$
which is equal to some KLM  entailment $C_{KLM}$ iff 
the MAK entailment $C_{MAK}$ is a
pre-circumscription.
\end{theo}
\underline{Proof:} 
The condition is necessary from Property \ref{propklmprecircct}. 
MAK entailments satisfy (CT) from Property 
\ref{propmakct}, thus Property \ref{propklmcar} gives the result.

This shows that what lacks to MAK 
entailment in order to be a KLM entailment is exactly the full preservation
of 
logical equivalence. 

Even if this result is satisfactory from a formal perspective, it is
more a characterization of the subclass of the MAK 
entailments which can be turned into KLM entailments than a characterization
of the subclass of MAK models which can be turned into KLM models. 
In some way, this makes no difference since 
the preferential entailments are fully defined from the models, but
we could expect an easier and more direct property, which can be checked
directly on the MAK model, without needing to compute its related
MAK entailment.

\subsection{A SUBCLASS OF MAK MODELS WHICH ARE KLM MODELS}
The fact that, for any KLM  entailment, 
we have $Th(\T) \incl C_{KLM}(\T)$ for any $\T\incl \La$ has already been
taken into account in \cite{DM92} for describing a subclass of the MAK models
which (1) can be turned into KLM models and (2) is powerful enough to 
give rise to all the MAK entailments which are also
KLM entailments. However, the description of the
subclass given in \cite{DM92} is needlessly complex. 
We describe here a simpler and more general subclass, which in our
opinion describes all the interesting and non pathological MAK models which
satisfy conditions (1) and (2) above.

\begin{defis}\label{defsupclass}
\begin{enumerate}
\item
A MAK model $\SS=(S, \trimod, \prec)$ is {\em supra classical}
whenever we get $Th(Cn_{\trimodi}(s))
=Cn_{\trimodi}(s)$ for any state $s$ in $S$.
This means that the ``world'' associated to each state is 
classically deductively closed (i.e. is a classical theory).
\item
An entailment $C$ is {\em supra classical}
if it satisfies 
$Th(\T) \incl C(\T)$ for any $\T\incl \La$.
\end{enumerate}
\end{defis}

The first definition concerns only the couple $(S, \trimod)$
describing the semantics to which $\prec$ will be applied, but is independent
of the preference relation $\prec$. Voorbraak \cite{Voo93P} calls
``$\La$-faithful'' what we call supra classical models.

For the second definition, notice that any pre-circum-scription is a supra
classical entailment.

\begin{lem}\label{lemmakthcs}
If a MAK model $\SS=(S, \trimod, \prec)$ is such that the entailment
$Cn_{\trimodi}$ is supra classical, then the MAK model $\SS$ is
supra classical.
\end{lem}
\underline{Proof:}
(1) $Cn_{\trimodi}(s) \incl Th(Cn_{\trimodi}(s))$ since 
$\T \incl Th(\T)$.

(2) $Th(Cn_{\trimodi}(s))\incl Cn_{\trimodi}(Cn_{\trimodi}(s))$ by hypothesis,
and we already know that we have in any case 
$Cn_{\trimodi}(Cn_{\trimodi}(s))=Cn_{\trimodi}(s)$,
which establishes $Th(Cn_{\trimodi}(s))=Cn_{\trimodi}(s)$. 

\begin{lem}\label{lemmakreci}
If a MAK model $\SS=(S, \trimod, \prec)$ is supra classical,
then the (Tarski) entailment $Cn_{\trimodi}$ and the (preferential) 
MAK entailment $C_{MAK}$ defined by $\SS$ are pre-circumscriptions.
\end{lem}

\underline{Proof:} 
(1) Let $\varphi \in Th(\T)$ and $s \in S(\T)$, i.e. 
$\T \incl Cn_{\trimodi}(s)$, then
$\varphi \in Cn_{\trimodi}(s)$ by supra classicality of $\SS$.
Thus, $\varphi \in \bigcap_{s\in S(\Ti)} Cn_{\trimodi}(s)$, i.e.
$\varphi \in Cn_{\trimodi}(\T)$: $Cn_{\trimodi}$ is supra classical.
From $Cn_{\trimodi}(\T) = \bigcap_{s\in S(\Ti)} Cn_{\trimodi}(s)$
we get also $Cn_{\trimodi}(\T)\in\TT$ since each $Cn_{\trimodi}(s)$ is in
$\TT$. 
Since by Definition \ref{defmakmod} we get
$C_{MAK}(\T) = \bigcap_{s\in S_{\prec}(\Ti)} Cn_{\trimodi}(s)$ 
(where the subset $S_{\prec}(\T)$ of $S(\T)$
is defined exactly as in Definition \ref{defklmmod} for KLM models), 
we get a fortiori $\varphi \in C_{MAK}(\T)\in\TT$.

(2) If $\T_1$ and $\T_2$ are two equivalent sets, 
then, by supra classicality of $\SS$, 
for each $s \in S$, we have $s\trimod \T_1$ iff
$s\trimod \T_2$, i.e. we have $S(\T_1) = S(\T_2)$.
A fortiori we get then $S_\prec(\T_1) = S_\prec(\T_2)$.
Thus we get $Cn_{\trimodi}(\T_1) = Cn_{\trimodi}(\T_2)$ and 
$C_{MAK}(\T_1)= C_{MAK}(\T_2)$ by the definitions of 
$Cn_{\trimodi}(\T)$ and $C_{MAK}(\T)$ respectively.

We have established:
\begin{prop}\label{propmaksupra}
\begin{enumerate}
\item
A MAK model \mbox{$\SS\!=\!(S,\trimod,\prec\!)$} is supra classical iff
the (Tarski) entailment $Cn_{\trimodi}$ that it defines is supra classical, iff
the entailment $Cn_{\trimodi}$ is a pre-circumscription.
\item If a MAK model is supra classical, then the (preferential) 
  MAK entailment $C_{MAK}$ that it determines is a
  pre-circumscription.  
\end{enumerate}
\end{prop}
We are now in position to establish our second main result:

\begin{theo}\label{thmaksupraklm}
\begin{enumerate}
\item If a MAK model $\SS$ is supra classical, then the MAK 
  entailment $C_{MAK}$  that it determines is equal to some KLM entailment
  $C_{KLM}$.

Precisely, if $\SS=(S,\trimod,\prec)$ is supra classical, 
then there exists a KLM model
\mbox{$\SS'=(S,l,\prec)$} with $S$ and $\prec$ unmodified, such that
the MAK entailment defined by $\SS$ is the KLM entailment
defined by $\SS'$.

\item Any KLM preferential entailment $C_{KLM}$ is equal to a MAK
  entailment defined by a supra classical MAK model.

Precisely, if $\SS=(S,l,\prec)$, then there exists
a supra classical MAK model \mbox{$\SS'=(S,\trimod,\prec)$}
with $S$ and $\prec$ unmodified, such that the KLM entailment
defined by $\SS$ is the MAK entailment defined by $\SS'$.
\end{enumerate}
\end{theo}
\underline{Proof:}
(1) Theorem \ref{thmakprecirc}
and Property \ref{propmaksupra}-2 give the first sentence.

Let $\SS=(S,\trimod,\prec)$ be a supra classical MAK model.
We get a KLM model as follows: we keep the set $S$ and the
relation $\prec$ unmodified. We define $l$ as the mapping 
$S\rightarrow \TT$ by taking $l(s)= Cn_{\trimodi}(s)$. 
It is immediate to see that
the KLM entailment $C_{KLM}$ is equal to $C_{MAK}$.

(2) We have a constructive proof already:
It suffices to see the construction given in 
Property \ref{propklmmmak}: it is clear from the definitions that
the MAK model obtained there is supra classical since we have already
noticed that each $l(s)$ in Definition \ref{defklmmod} can be equated to a 
classical theory. Notice that this theorem could also have been obtained as 
a consequence of some results in an earlier independent work by
Voorbraak \cite{Voo93P}. Rather strangely, 
Voorbraak does not enounce this result in all generality, referring to
\cite{DM92} for further results on the subject.\\ \ 

Thus, we get characterization results and constructive passages 
simpler and easier than those given in \cite{DM92}. However, our results
are slightly more general (see why in note \ref{not3b}): 
the subclass of the MAK models considered here is
slightly greater than the subclass considered 
by Dix and Makinson since they consider a strict subclass of the MAK models 
which can be ``amplified'' (in their terms).  It is easy to see, from
\cite{DM92} together with our results,  that the class of the MAK models 
which can be ``amplified'' coincide with the class of the supra classical 
MAK models.   
Moreover, our comparison does not need to consider a third intermediate
(between $Cn_{\trimodi}$ and $C_{MAK}$) non classical entailment%
\footnote{\label{not3a}%
For readers familiar with \cite{DM92}, let us notice that an immediate
consequence 
of our results is that, even in the exact framework and formulation considered
in \cite[main theorem]{DM92}, the condition (3a) given there 
($s\trimod \varphi$ and $s\trimod \varphi'$ implies 
 $s\trimod\varphi\et\varphi'$) is redundant.},
which plays an important role in the results of \cite{DM92}, but which 
complicates the  direct comparison between KLM and MAK preferential
entailments. 
This simplification comes mainly from our results about supra classical MAK
models. And the condition that each $Cn_{\trimodi}(s)$ must be a theory is
easily checked, without the need to compute the associated MAK entailment or
to introduce a third non classical entailment.\\ \ 

One consequence of our results about MAK entailments is that, if we 
are concerned only by those MAK entailments which respect full logical
equivalence, then we can restrict our attention to a yet narrower class
of MAK models. Indeed, we have seen just above why in this case we can
restrict our attention 
to the easily described class of supra classical MAK models. Now, since we
know that, for KLM entailments, we can consider only the simplified version
of KLM models,
our passages between MAK models and KLM models show that we can 
also require a
{\em unicity of states} condition for MAK models.
By ``unicity of states'', we mean that, for any
different  $s,s'\in S$, the ``worlds'' $Cn_{\trimodi}(s)$ and
$Cn_{\trimodi}(s')$ corresponding to these two states are different.
The class of the supra classical MAK models satisfying unicity of states is
powerful enough to generate all the MAK entailments which are
pre-circumscriptions. 
Let us describe briefly
now the analogous of the singular and the strictly singular KLM models in
terms of MAK models. 

\begin{defi}\label{defmakclas}
A MAK model  $\SS=(S, \trimod, \prec)$ 
is {\em classical} if the ``worlds'' $Cn_{\trimodi}(s)$
are (classical) complete theories, for any state $s$ in $S$.
\end{defi}
\begin{rem}\label{remclas}
$\bullet$
A MAK model is supra classical iff the satisfaction relation $\trimod$
respects the binary connector $\et$: for each $s\in S$, we have
\[s\trimod \varphi_1 \et \varphi_2 \makebox[3em]{iff}
s \trimod \varphi_1 \makebox[2em]{and} 
s \trimod \varphi_2.\hspace{1em}(R_\et)\] 
$\bullet$ $\bullet$
A MAK model is classical iff it is supra classical and $\trimod$ respects the
negation $\non$:  
\[s |\!\!\!\not \equiv \varphi \makebox[3em]{iff} s \trimod
\non\varphi.\hspace{1em} (R_\non)\] \end{rem}

\underline{Proof:}
$\bullet$ If each $Cn_{\trimodi}(s)$ is in
$\TT$, then, since $\{\varphi_1, \varphi_2\}\equiv \{\varphi_1\et\varphi_2\}$, 
we get $\{\varphi_1, \varphi_2\}\incl Cn_{\trimodi}(s)$ iff 
$\{\varphi_1\et\varphi_2\}\incl Cn_{\trimodi}(s)$. Conversely, let us suppose 
$(R_\et)$. Then, if $\{\varphi_i\}_{\in I}\incl Cn_{\trimodi}(s)$ and
$\{\varphi_i\}_{i\in I} \mod \varphi$, by compactness of $\mod$ there exists
a finite $J\incl I$ such that $\{\varphi_i\}_{i\in J} \mod \varphi$, i.e.
$\bigwedge_{i\in J}\varphi_i \mod \varphi$, i.e.
$\bigwedge_{i\in J}\varphi_i \equiv (\bigwedge_{i\in J}\varphi_i) \et
\varphi$, thus, by $(R_\et)$, $\varphi \in Cn_{\trimodi}(s)$.
Remind that we identify a formula with its equivalence class.
Makinson does not always make this assumption in \cite{Mak94G}, thus,
his original formalism is slightly   
more general than the version given in the present text. However, since 
with KLM entailments a formula can always be replaced by an equivalent 
formula, we have to make this
assumption (or any equivalent one) when we want to compare the two formalisms.
This means that if this assumption is not made till the beginning (as in this
text), then it must be added, e.g. by requiring in Definition
\ref{defsupclass}-1 that $\mod$ is standard. Notice that Definition
\ref{defsupclass}-1 as it stands 
implies that two formulas equivalent (for $\mod$) are
always in the same sets $Cn_{\trimodi}(s)$, thus are ``equivalent for
$\trimod$''.  

$\bullet$ $\bullet$ A theory $Cn_{\trimodi}(s)$ is complete iff 
[$\varphi \in Cn_{\trimodi}(s)$ iff $\non\varphi\notin Cn_{\trimodi}(s)$],
i.e. iff
 $\trimod$ satisfies $(R_\non)$.

A MAK model is classical iff $\trimod$ respects all the logical
connectors. For instance, it is immediate to see that $(R_\et)$ and $(R_\non)$
imply $(R_\ou)$:
\[s\trimod \varphi_1 \ou \varphi_2 \makebox[3em]{iff}
s \trimod \varphi_1 \makebox[2em]{or} 
s \trimod \varphi_2.\hspace{1em}(R_\ou)\] 

Classical MAK models correspond to singular KLM models while
the  classical  MAK models which respect unicity of states
correspond to the  strictly singular KLM models.

\subsection{COMING BACK TO THE ORIGINAL KLM ENTAILMENTS}
This work applies also to cases where special conditions are
required for the models. We think that the simplicity and the
naturalness of our translation is a first serious indication for this.
Let us consider the original definitions.
\begin{defis}\label{defsmooth}
  \begin{enumerate}
  \item A {\em consistent KLM model}, is such that each
state is {\em consistent}, meaning that $l(s)$ is consistent.
\item 
A KLM model $\SS=(S,l,\prec)$ is
{\em smooth} ({\em stoppered} in \cite{Mak94G}) if, for each $\T\incl\La$
and $s\in S(\T)-S_\prec(\T)$, there exists $s'\in S_\prec(\T)$ 
such that $s'\prec s$ (``minoration by a minimal state'').
  \end{enumerate}
\end{defis}
\cite{KLM90} considers only the KLM models which are consistent and smooth.
The authors consider that the ``converse of (CT)''
[if $\T' \incl C(\T)$, then $C(\T) \incl C(\T \union \T')$],
called {\em cumulative monotony} (CM), is as important as (CT), and they only
care of {\em cumulative entailments}, which satisfy (CT) and (CM).
They give the following characterization:\\ \ 

{\bf Original KLM characterization} \cite{KLM90}:\\
A pre-circumscription $C$ is cumulative iff it is a KLM entailment  
defined by a smooth and consistent KLM model.\\ \ 

It happens that this characterization result also holds without the
consistency condition, which confirms our opinion made after Definition
\ref{defklmmod} that, for KLM entailments, the requirement that $l(s)$ must
be consistent is needless\footnote{\label{not3b}%
For readers familiar with \cite{DM92}, let us remind that Dix and Makinson add
to (3a) (see note \ref{not3a}) a ``consistency of states'' condition
(3b), for the MAK models, in their main theorem. This condition is necessary
in their text only because they disallow inconsistent states in KLM models,
following \cite{KLM90}.  Since inconsistent 
states are not a real problem, this condition, which restricts slightly the
class of the MAK models concerned,  
can be suppressed without modifying the
results about the preferential 
entailments.}.

We get, with the KLM models as defined here, a first
modification of KLM characterization:\\ \ 

{\bf A ``KLM Characterization'' allowing inconsistent states:} 
A pre-circumscription $C$ is cumulative iff it is a KLM entailment 
defined by a smooth KLM model.\\ \ 

 The proof is an easy modification of  the proof of
the original characterization \cite{KLM90}, moreover this result has already
appeared as \cite[Observation 3.4.5]{Mak94G} and \cite[Proposition
5.4]{Voo93P}. 

We can go even further, by requiring that the KLM model is a {\em
  simplified KLM model}, meaning an ``unlabelled model'', or a 
``model without states'': 
$\;S=\TT\;$ and $\;l = identity$.\\ \ 

{\bf A ``KLM Characterization'' with simplified models:} \\
A pre circumscription $C$ is cumulative iff it is a KLM entailment
defined by a smooth simplified KLM model.\\ \  

Indeed, the ``if'' side comes from the ``if'' side of the previous 
characterization, allowing inconsistent states.
For the
``only if'' side, it suffices to define the simplified KLM model associated
to $C$ as follows:

(1) $S= \TT, l = identity$ (the model is simplified, no need for states).

(2) For  $\T_1, \T_2$ in $\TT$, $\T_1 \prec_C \T_2$ iff

(2a) $\T_1=\La$ and $\T_2 \not = f(\T)$ for any $\T\in\TT$, or

(2b) $\T_2 \not =\La$,  $\T_2 \not =\T_1$ and there exists $\T_3,\T_4$\\
\makebox[2em]{}
 in $\TT$ such that $f(\T_3) = \T_1$, $f(\T_4) = \T_2$,\\
\makebox[2em]{} and $\T_3 \incl \T_2$.

Then, if $C$ is a cumulative pre-circumscription, an easy translation of 
the proof of the characterization result from \cite{KLM90}
(where $l$ is injective, as also taken into account in a ``suppression of
states'' result given in \cite{Boc99U}) shows that we get, in a way very
similar to the original proof of \cite{KLM90}: 

(1) $C$ is equal to the KLM entailment defined by\\
\makebox[1em]{} this simplified KLM
  model, and

(2) this [simplified] KLM model is smooth \\
\makebox[1em]{} (and irreflexive). 

Notice that this result has also appeared as \cite[Proposition 5.5]{Voo93P},
with an apparently different proof.

We can then get immediately the corresponding characterization results in
terms of MAK entailment, 
by using Theorems \ref{thmakprecirc} and \ref{thmaksupraklm}.

\section{CONCLUSION AND PERSPECTIVES}
We have shown that the notions of preferential entailment
as defined by Kraus, Lehmann and Magidor 
and as defined by Makinson are much closely related than was
supposed before. Indeed, these two notions coincide exactly in all the 
cases where they can coincide, that is when the underlying logic 
respects classical equivalence.
Moreover, we have shown that a similar result holds also for the
respective models defining the two notions.
It was already known that any KLM model could easily be turned into a MAK
model. We have exhibited a natural subclass of the MAK models which can,
exactly as easily, be turned into a KLM model. The subclass obtained here is 
slightly greater, and is much easier to describe, than what was previously
known.  And this subclass of models is ``complete'': it generates all the KLM
preferential entailments.
This subclass is the class of the MAK models for which all the states have a
``classical'' behavior: the set of formulas they satisfy is closed for
classical deduction.
This subclass is the most natural class to consider.
Indeed, this is the class such that, for any preference relation $\prec$,
we are certain from the beginning that the MAK preferential entailment
generated has a classical behavior with respect to logical equivalence.
There exist some MAK models outside this class which give rise to a 
KLM entailment, but these models are rather special, since it turns out
that their preference relation, in some way, 
eliminates all the states in the model with an
unclassical behavior.
We have also shown that our results apply to important particular 
subclasses
of KLM models and MAK models, namely those which are simplified in that
either 
the labelling mapping $l$ is needless [KLM side], 
or some ``unicity of state'' condition is required [MAK side].
And we have shown that, even for the cumulative entailments considered in the
original texts, these simplified models suffice, and that the passages between 
KLM models and MAK models work in this case also.
As (non trivial) future work, let us remark that
these results should help further study on the subject, since they show
that this kind of preferential entailment is not as ``cumbersome'' as 
it is qualified even in the founding paper \cite{KLM90}.
Even automatic computation could take advantage from these results, 
since the  models
considered here have nice properties, which, hopefully, could help
designing new kinds of ``preferential entailments demonstrators''.

\subsubsection*{Acknowledgements}
The author wants to thank the referees for their very useful comments.

\end{document}